\documentclass[lettersize,journal]{IEEEtran}
\usepackage{amsmath,amsfonts}
\usepackage{array}
\usepackage[caption=false,font=normalsize,labelfont=sf,textfont=sf]{subfig}
\usepackage{textcomp}
\usepackage{stfloats}
\usepackage{url}
\usepackage{verbatim}
\usepackage{graphicx}
\usepackage{cite}
\usepackage{multirow}
\usepackage{booktabs}
\usepackage{bbding}
\usepackage{pifont}
\usepackage{listings}

\begin{document}

\title{BjTT: A Large-scale Multimodal Dataset for\\ Traffic Prediction}

\author{Chengyang Zhang, Yong Zhang\thanks{Yong Zhang is the corresponding author.}, Qitan Shao, Jiangtao Feng, Bo Li, Yisheng Lv, Xinglin Piao and Baocai Yin

\thanks{Chengyang Zhang, Yong Zhang, Qitan Shao, Jiangtao Feng, Bo Li, Xinglin Piao and Baocai Yin are with Beijing Key Laboratory of Multimedia and Intelligent Software Technology, Beijing Artificial Intelligence Institute, Faculty of Information Technology, Beijing University of Technology, Beijing, China, 100124. (Cy\_Zhang@emails.bjut.edu.cn; zhangyong2010@bjut.edu.cn; shaoqt@emails.bjut.edu.cn; fengjt129@emails.bjut.edu.cn; bo\_li@emails.bjut\\.edu.cn; piaoxl@bjut.edu.cn; ybc@bjut.edu.cn)}
\thanks{Yisheng Lv is with Institute of Automation, Chinese Academy of Sciences. (yisheng.lv@ia.ac.cn)}
\thanks{The research project is partially supported by the National Key R\&D Program of China (No. 2021ZD0111902), National Natural Science Foundation of China (No.62072015, U21B2038, U19B2039, 61902053) and Beijing Natural Science Foundation (4222021).}
}

\markboth{Journal of \LaTeX\ Class Files,~Vol.~14, No.~8, August~2021}%
{Shell \MakeLowercase{\textit{et al.}}: A Sample Article Using IEEEtran.cls for IEEE Journals}


\maketitle

\begin{abstract}
Traffic prediction plays a significant role in Intelligent Transportation Systems (ITS). Although many datasets have been introduced to support the study of traffic prediction, most of them only provide time-series traffic data. However, urban transportation systems are always susceptible to various factors, including unusual weather and traffic accidents. Therefore, relying solely on historical data for traffic prediction greatly limits the accuracy of the prediction. In this paper, we introduce Beijing Text-Traffic (BjTT), a large-scale multimodal dataset for traffic prediction. BjTT comprises over 32,000 time-series traffic records, capturing velocity and congestion levels on more than 1,200 roads within the 5th ring area of Beijing. Meanwhile, each piece of traffic data is coupled with a text describing the traffic system (including time, location, and events). We detail the data collection and processing procedures and present a statistical analysis of the BjTT dataset. Furthermore, we conduct comprehensive experiments on the dataset with state-of-the-art traffic prediction methods and text-guided generative models, which reveal the unique characteristics of the BjTT. The dataset is available at \url{https://github.com/ChyaZhang/BjTT}.
\end{abstract}

\begin{IEEEkeywords}
Traffic prediction, large-scale, new dataset.
\end{IEEEkeywords}

\section{Introduction}
\label{section1}
\IEEEPARstart{T}{raffic} prediction stands as a prominent and vital task within the Intelligent Transportation Systems (ITS). The precise and timely prediction of traffic situations plays a significant role in refining individual travel decisions and alleviating traffic congestion. With increasing attention from researchers on traffic prediction, numerous datasets have been introduced to support the development of this field. Since traffic prediction relies heavily on historical data collected from road sensors, the majority of existing datasets \cite{1,2,3,4} focus on providing more historical traffic data.

In fact, the urban transportation system is inherently intricate and subject to various influencing factors like unusual weather, road traffic accidents, and large social events. Relying solely on historical traffic data proves challenging in capturing these concealed factors. Therefore, it is crucial to incorporate these factors for a more accurate and realistic prediction of traffic situations. In recent years, there have been many advanced traffic prediction methods appearing. While traditional machine learning methods \cite{7,8,9,10,11} were prevalent earlier, contemporary methods predominantly leverage deep learning techniques, particularly Long Short-Term Memory (LSTM) \cite{12,13,14,15,16} and Graph Neural Networks (GNN) \cite{3,4,5,6,17,18,19}. Despite the success in short-term predictions, there are still two main challenges faced by these methods: 1) Poor performance in long-term prediction. 2) Insensitive to abnormal events. However, optimizing temporal prediction methods alone with existing insufficient datasets falls short of overcoming these two challenges. 

Over the past decades, a lot of datasets have been benchmarked for traffic prediction studies. The data sourced from the Caltrans Performance Measurement System (PeMS) \cite{1} is almost the most popular open-source dataset. However, direct utilization of PeMS data for traffic prediction tasks is not feasible and requires preprocessing. Consequently, the selection of observation points and their quantity in PeMS varies across different works. For example, commonly used PeMS series datasets include PeMS03, PeMS04, PeMS07, PeMS08, PeMSD7-M, PeMSD7-L, METR-LA and PeMS-BAY \cite{2,26}. In addition, several works evaluate methods on other datasets. Typically, Yu et al. \cite{3} evaluate their method on BJER4, which originated from the main areas of east ring No.4 routes in Beijing. Guo et al. \cite{4} collect the traffic data of Jinan and Xian from the Didi Chuxing GAIA Initiative. However, these data are limited in certain aspects, including the number of sensors, time range, data accessibility, and data type. To this end, there is a compelling need to introduce a new open-source dataset characterized by a more comprehensive volume and variety of data types for prompting traffic prediction research.

\begin{figure*}[t!]
  \centering
   \includegraphics[width=2.\columnwidth]{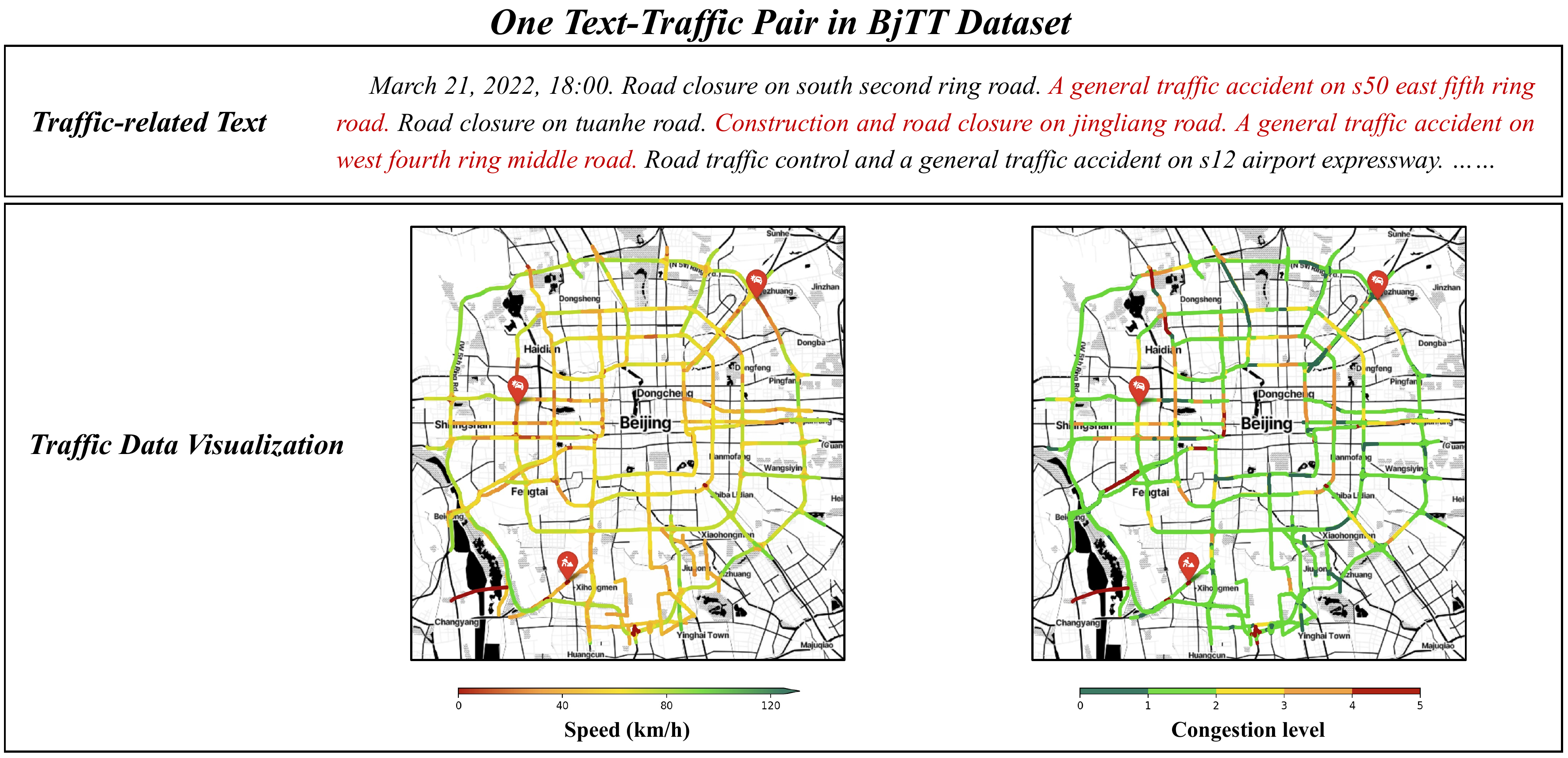}
   \caption{The illustration of one text-traffic pair within
the BjTT dataset. Each traffic data contains different types of road information and is coupled with a text describing the traffic system.}
   \label{fig1}
\end{figure*}

In this paper, we introduce the Beijing Text-Traffic (BjTT), a substantial multimodal dataset tailored for traffic prediction. Figure \ref{fig1} provides an illustration of one text-traffic pair within the BjTT dataset. Compared with existing datasets, BjTT mainly encompasses the following properties:
\begin{itemize}
    \item \textbf{Large data volume:} BjTT includes more than 32,000 time-series traffic data spanning 3 months, each capturing information from more than 1,200 major roads within the fifth ring road area of Beijing. 
    
    \item \textbf{Diverse data types:} Each traffic data includes two distinct types of road information, which are velocity and congestion level, providing a comprehensive view of road situations. 

    \item \textbf{New data modality:} In BjTT, each traffic data is coupled with a traffic-related text description. The text contains the timestamp, date, and details of abnormal events that occurred on the road at that moment. The abnormal events cover a wide range, including road accidents, construction activities, weather conditions, and more than 30 other event types. To the best of our knowledge, BjTT stands out as the most extensive open-source dataset that integrates both traffic and event data.
\end{itemize}
The comparative overview of the BjTT dataset against existing mainstream traffic prediction datasets is listed in Table \ref{table1}, which highlights the BjTT as a comprehensive multimodal dataset. We hope our BjTT dataset can enable researchers to consider more challenging and practical problems in the field of traffic prediction.

\begin{table*}[t]
\renewcommand{\arraystretch}{1.3}
\caption{Comparisons of our BjTT dataset with existing traffic prediction datasets. BjTT achieves leadership in both data volume and data type. In addition, BjTT offers the text that describes events on the road as a traffic prediction dataset for the first time.}
\centering
\resizebox{\textwidth}{!}
{\begin{tabular}{c|c|c|c|c|c|c|c|c|c}
\toprule[1pt]
                  & \textbf{PeMS03}                                              & \textbf{PeMS04}                                              & \textbf{PeMS07}                                              & \textbf{PeMS08}                                              & \textbf{PeMSD7(M)}                                           & \textbf{PeMSD7(L)}                                             & \textbf{PeMS-BAY}                                            & \textbf{METR-LA}                                             & \textbf{BjTT}                                                               \\ \hline
\textbf{$|\mathcal{V}|$}                 & 358                                                          & 307                                                          & 883                                                          & 170                                                          & 228                                                          & 1,026                                                        & 325                                                          & 207                                                          & 1,260                                                                       \\ \hline
\textbf{Time steps}        & 26,208                                                       & 16,992                                                       & 28,224                                                       & 17,856                                                       & 12,672                                                       & 12,672                                                       & 52,116                                                       & 23,272                                                       & 32,400                                                                      \\ \hline
\textbf{Time range}        & \begin{tabular}[c]{@{}c@{}}09/2018 - \\ 11/2018\end{tabular} & \begin{tabular}[c]{@{}c@{}}01/2018 - \\ 02/2018\end{tabular} & \begin{tabular}[c]{@{}c@{}}05/2017 - \\ 08/2017\end{tabular} & \begin{tabular}[c]{@{}c@{}}07/2016 - \\ 08/2016\end{tabular} & \begin{tabular}[c]{@{}c@{}}05/2012 - \\ 06/2012\end{tabular} & \begin{tabular}[c]{@{}c@{}}05/2012 - \\ 06/2012\end{tabular} & \begin{tabular}[c]{@{}c@{}}01/2017 - \\ 05/2017\end{tabular} & \begin{tabular}[c]{@{}c@{}}03/2012 - \\ 06/2012\end{tabular} & \begin{tabular}[c]{@{}c@{}}01/2022 - \\ 03/2022\end{tabular}                \\ \hline
\textbf{Traffic data type} & Volume                                                       & Volume                                                       & Volume                                                       & Volume                                                       & Velocity                                                     & Velocity                                                     & Velocity                                                     & Velocity                                                     & \begin{tabular}[c]{@{}c@{}}Velocity, \\Congestion level\end{tabular} \\ \hline
\textbf{Textual event data}        & \ding{55}                                                            & \ding{55}                                                            & \ding{55}                                                            & \ding{55}                                                            & \ding{55}                                                            & \ding{55}                                                            & \ding{55}                                                            & \ding{55}                                                            & \ding{51}                                                                           \\ \bottomrule[1pt]
\end{tabular}}
\label{table1}
\end{table*}

Our major contributions are summarized as follows:
\begin{itemize}
\item We release the BjTT, the first publicly multimodal dataset containing traffic data and event descriptions for traffic prediction. With over 32,000 text-traffic pairs, BjTT is more practical than existing datasets. 

\item We evaluate several state-of-the-art traffic prediction methods on the BjTT dataset and analyze their performance. The experimental results can help us to analyze the traffic prediction performance of existing methods more comprehensively and fairly.

\item We involve a text-guided generative model in the experiments to demonstrate how multimodal data can assist traffic prediction. The results reveal that text-guided traffic prediction can help alleviate the poor long-term prediction performance of existing methods. 
\end{itemize}

The structure of the rest paper is organized as follows. Section \ref{section2} explores the discussion on related works. Section \ref{section3} offers insights into the construction and statistical properties of the dataset. Experimental results and analysis are detailed in Section \ref{section4}. Finally, Section \ref{section5} covers the conclusion and future perspectives.

\section{Related Works}
\label{section2}
\subsection{Traffic Prediction Methods}
Traffic prediction plays a crucial role in assisting governments to effectively manage the urban transportation system. Presently, traffic prediction methods can be classified into two categories: classical methods and deep learning methods, while classical methods include parameter methods and non-parameter methods. Parameter methods are characterized by a fixed model structure, and their parameters can be computed via empirical data. In the 1970s, Ahmed \textit{et al.} \cite{20} propose the Auto-Regressive Integrated Moving Average (ARIMA) to predict the short-term highway traffic flow. In the next decades, the variants of ARIMA \cite{21,22,23} have emerged as some of the most consolidated methods based on classical statistics. However, these methods are constrained by the stationary assumption of time sequences and do not consider spatial-temporal correlations. Non-parameter methods have no fixed structures and parameters, which can fit nearly all the functions to arbitrary precision. For instance, Vanajakshi \textit{et al.} \cite{24} examine the use of Support Vector Machines (SVM) for the short-term prediction of travel time. Jeong \textit{et al.} \cite{25} present the Online Learning Weighted Support-Vector Regression (OLWSVR) to incorporate the temporal differences in traffic flow data.

As urban transportation systems continue to evolve, classical methods are no longer sufficient to handle the massive and intricate traffic data. Therefore, deep learning methods have gathered much attention and have been widely applied to various traffic prediction works. Typically, Lv \textit{et al.} \cite{14} first apply the Stacked AutoEncoder (SAE) to learn generic traffic flow features and achieve superior performance. Shi \textit{et al.} \cite{16} propose the convolutional LSTM to build an end-to-end trainable model for precipitation nowcasting, which is an extended fully connected LSTM. Fu \textit{et al.} \cite{15} associate the LSTM and Gated Recurrent Unit (GRU) to predict the short-term traffic flow. However, the traditional convolution restricts the model to only process grid structures rather than general domains. To better represent the irregular traffic network structure, researchers integrate Graph Convolutional Networks (GCN) into the traffic prediction to extract spatial correlations. Specifically, Yu \textit{et al.} \cite{3} innovatively propose the Spatio-Temporal Graph Convolutional Networks (STGCN), which combines GCN with gated causal convolution to address time series prediction challenges in the domain of traffic. Wu \textit{et al.} \cite{5} present an advanced graph neural network known as Graph WaveNet (GWN), specifically developed for the analysis of spatial-temporal graphs. This model introduces a novel approach by incorporating an adaptive dependency matrix, which is derived through node embedding techniques. This allows for the accurate discovery of concealed spatial relationships in the dataset. Choi \textit{et al.} \cite{27} introduce the method of Spatio-Temporal Graph Neural Controlled Differential Equation (STG-NCDE), which comprises two distinct NCDEs: one dedicated to temporal dynamics and the other to spatial analysis.

With the increasing appearance of complex traffic prediction models, the performance of time-series prediction is gradually plateauing. To address the current challenges in the traffic prediction task, solely relying on historical traffic data has become challenging. Consequently, a key research focus for the future lies in exploring ways to effectively incorporate additional traffic information into the traffic prediction task.

\subsection{Traffic Prediction Datasets}
Traffic prediction study has made significant progress along with the emergence of a large amount of datasets. Typically, the data sourced from the PeMS database serves as prevalent benchmarks for evaluating traffic prediction methods. However, the temporal scope and sensor stations in the selected data vary among studies. To enable a fair comparison of traffic prediction methods, some researchers have released a series of collected and processed PeMS datasets. For example, Song \textit{et al.} \cite{26} construct PeMS03, PeMS04, PeMS07 and PeMS08. The flow data is aggregated to 5 minutes in these four datasets, resulting in 12 data points per hour. It is worth noting that these four datasets primarily involve volume data, while subsequent datasets focus on velocity. Yu \textit{et al.} \cite{3} select a medium and a large range among the District 7 of California, containing 228 and 1,026 stations, labeled as PeMSD7(M) and PeMSD7(L), respectively. The PeMSD7 dataset covers weekdays in May and June of 2012. Additionally, Li \textit{et al.} \cite{2} select 207 sensors on the highway of Los Angeles County, collecting data for 4 months from March 1st, 2012, to June 30th, 2012, known as METR-LA. Besides, they also introduce the PeMS-BAY dataset, which contains 6 months of data from January 1st, 2017 to May 31st, 2017.

Table I provides an overview of the distinct properties of existing datasets, revealing various limitations in each. A critical observation is that all these datasets predominantly contain unimodal data. To propel advancements in the field of traffic prediction, it is crucial to construct multimodal datasets that encompass a more comprehensive set of road information.

\begin{figure*}[t]
  \centering
   \includegraphics[width=2.\columnwidth]{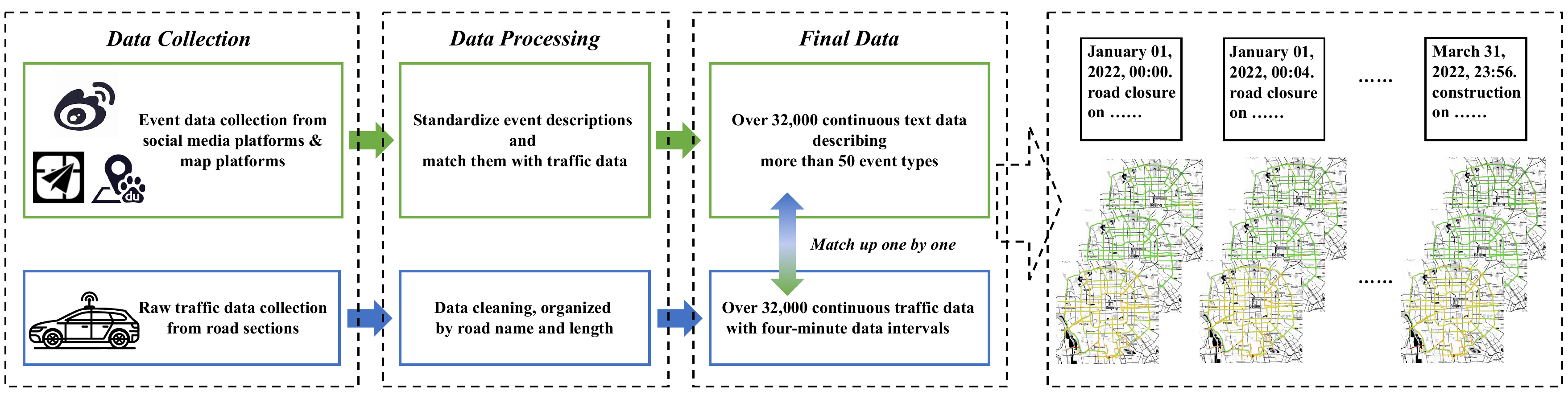}
   \caption{The construction pipeline of the BjTT dataset includes two main parts: data collection and data processing. The processed traffic (blue boxes) and text (green boxes) data are matched up one by one to get the final dataset.}
   \label{fig2}
\end{figure*}

\section{The BjTT dataset}
\label{section3}
We aim to provide a large-scale traffic prediction dataset with comprehensive data types and textual descriptions to expose the challenges of traffic prediction. In this section, we present the data construction of BjTT, followed by the analysis of dataset statistics and characteristics.

\subsection{Dataset Construction}
The dataset construction pipeline is illustrated in Figure \ref{fig2}. The pipeline mainly includes two parts: data collection and data processing. The blue and green boxes represent the traffic and event textual data construction procedures. Next, we will detail the data collection and data processing.

\subsubsection{Data collection}
To construct the BjTT dataset, we need to collect both traffic and event data. For the traffic data, the defined boundary for data collection is the fifth ring expressway area of Beijing. We collect the historical traffic data from some map service providers such as Amap and Baidu map. These traffic data span a period of three months and contain different types of information for each road section. For the event data, we collect event data from some social platforms such as Weibo and Little Red Book via the typical query keywords related to the traffic, e.g., traffic accidents, road construction, and so on. In addition, we also record road events data according to several map applications as a supplement. Finally, all event data undergoes a deduplication process to eliminate redundancy.

\begin{figure}[t]
  \centering
   \includegraphics[width=1.\columnwidth]{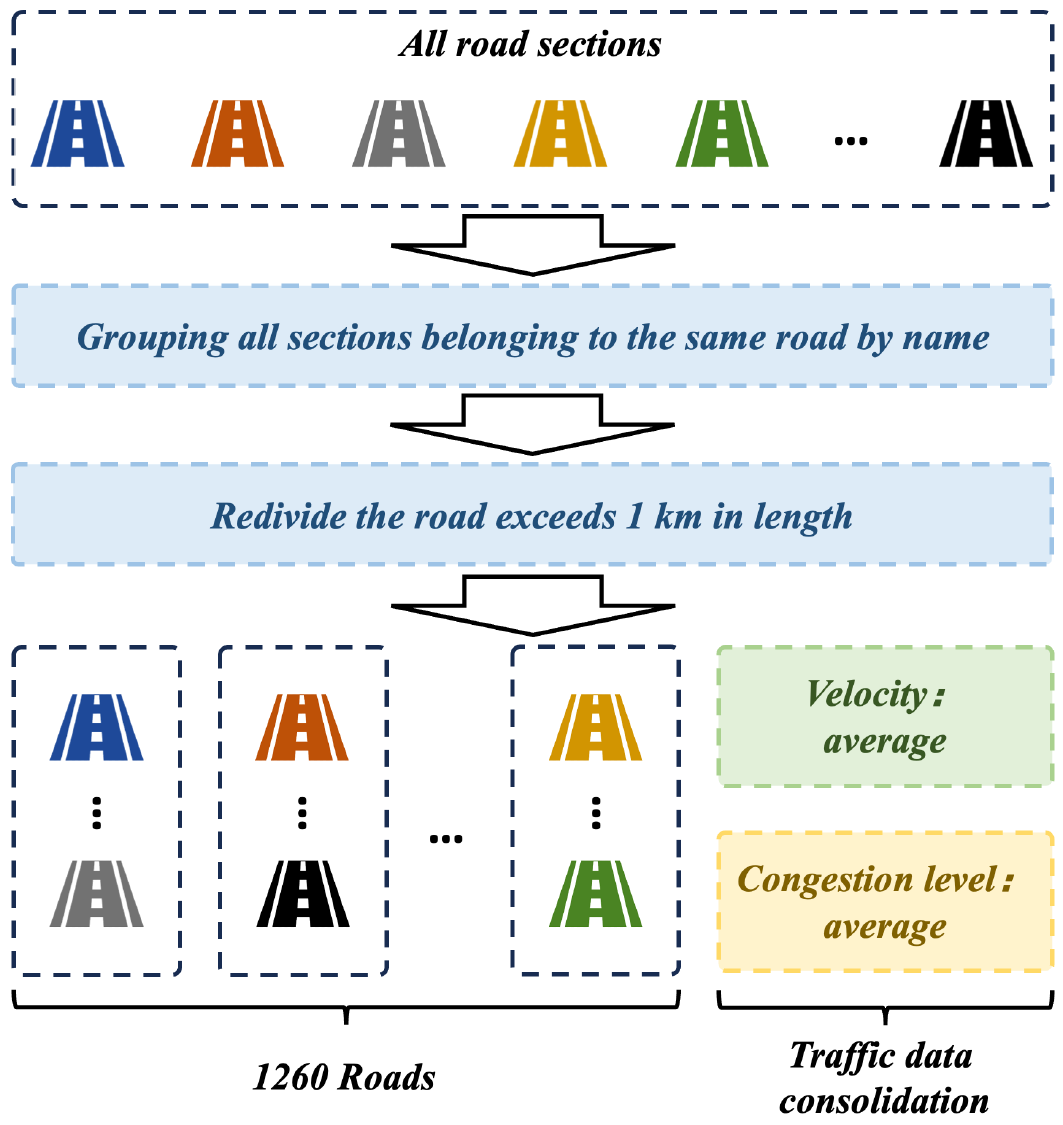}
   \caption{Data grouping process for road sections. Road sections are grouped by the road name, and the velocity and congestion level of the grouped roads are calculated by average.}
   \label{fig3}
\end{figure}

\begin{figure}[t]
  \centering
   \includegraphics[width=1.\columnwidth]{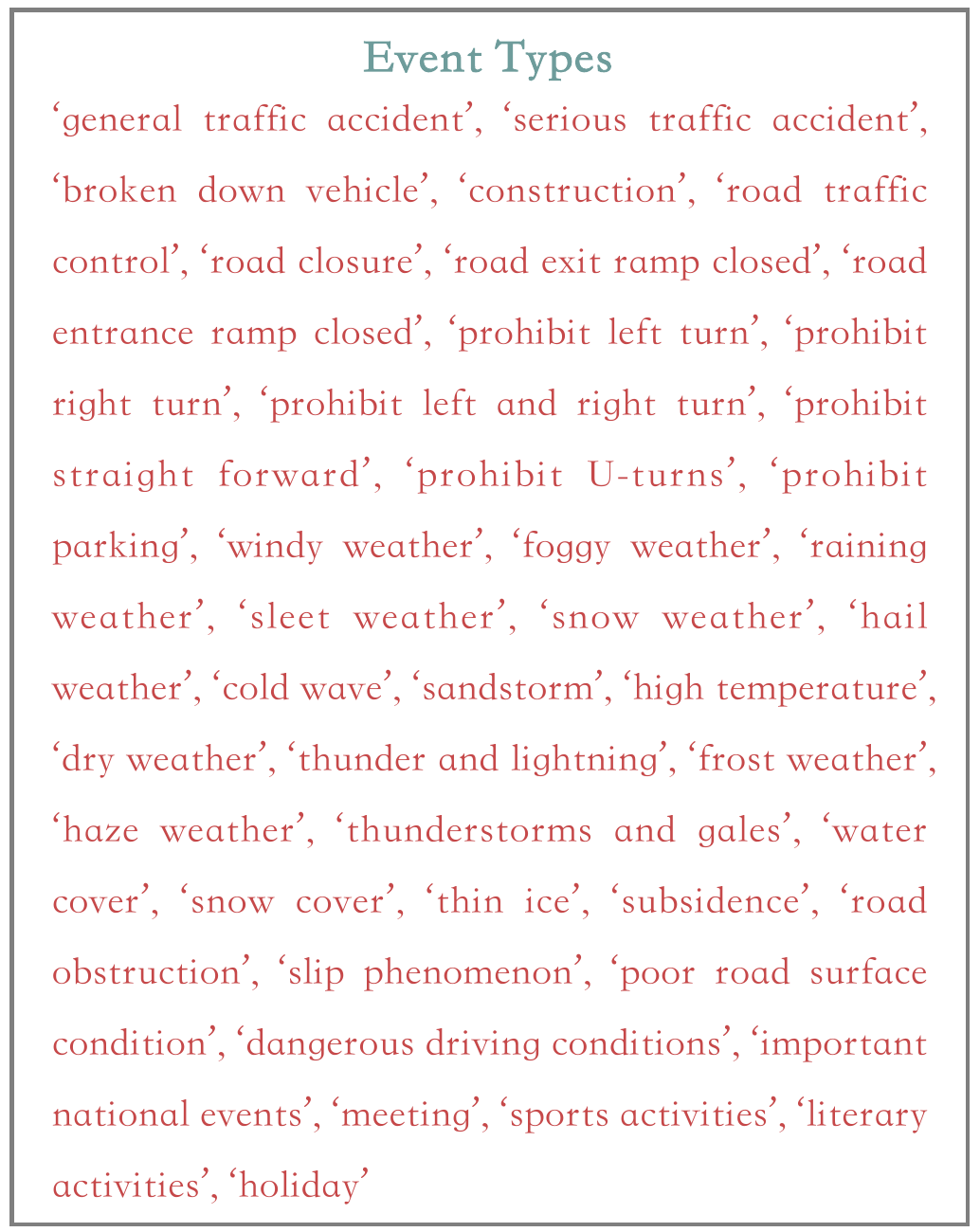}
   \caption{All event types that included in the BjTT dataset. These events cover not only common road traffic events such as construction and traffic accidents but also unusual weather and large social events.}
   \label{fig4}
\end{figure}

\subsubsection{Data processing}
First of all, we need to merge the raw traffic data. For a specific road, it is divided into several small road sections on the map, and each section stores traffic information separately. Similarly, the data we collect contains all or part of the road sections for each road in the delineated area. To avoid the effect of missing road sections, we group the information of road sections belonging to the same road, as shown in Figure \ref{fig3}. Specifically, we first group all road sections by the road name. If a grouped road exceeds 1 km in length, we divide it into two separate roads again. In this way, we have 1,260 roads that are less than 1 km in length, with some sharing the same road name. The final speed and congestion level for each of the 1,260 roads represent the averages across all road sections that make up the road. This method is applied to process traffic data at each time point, thus refining and consolidating the traffic data. Finally, we represent the traffic data at each time point in a matrix $x_{i}, x_{i}\in \mathbb{R}^{1260\times 2}, i=1, 2, ..., 32,400$.
To process event data, we initially identify event keywords from the raw data, as depicted in Figure \ref{fig4}. We extract over 40 diverse types of events, encompassing typical road traffic occurrences like construction and traffic accidents, as well as abnormal weather and significant social events. Subsequently, we organize the events that occur at each time point as text, following the format "event + location." To establish a one-to-one correspondence with the traffic data, date and time descriptions are prepended to all text, creating a comprehensive record, as illustrated in Figure \ref{fig1}.

\subsubsection{Copyright and privacy} 
BjTT does not hold the copyright of the traffic and event data. The copyright of these data belongs to the map service providers and social media platforms. Researchers accept full responsibility for using BjTT and only for non-commercial research and educational purposes. Furthermore, some road names in the event data are not entirely authentic due to confidentiality concerns.

\begin{figure*}[t]
  \centering
   \includegraphics[width=\textwidth]{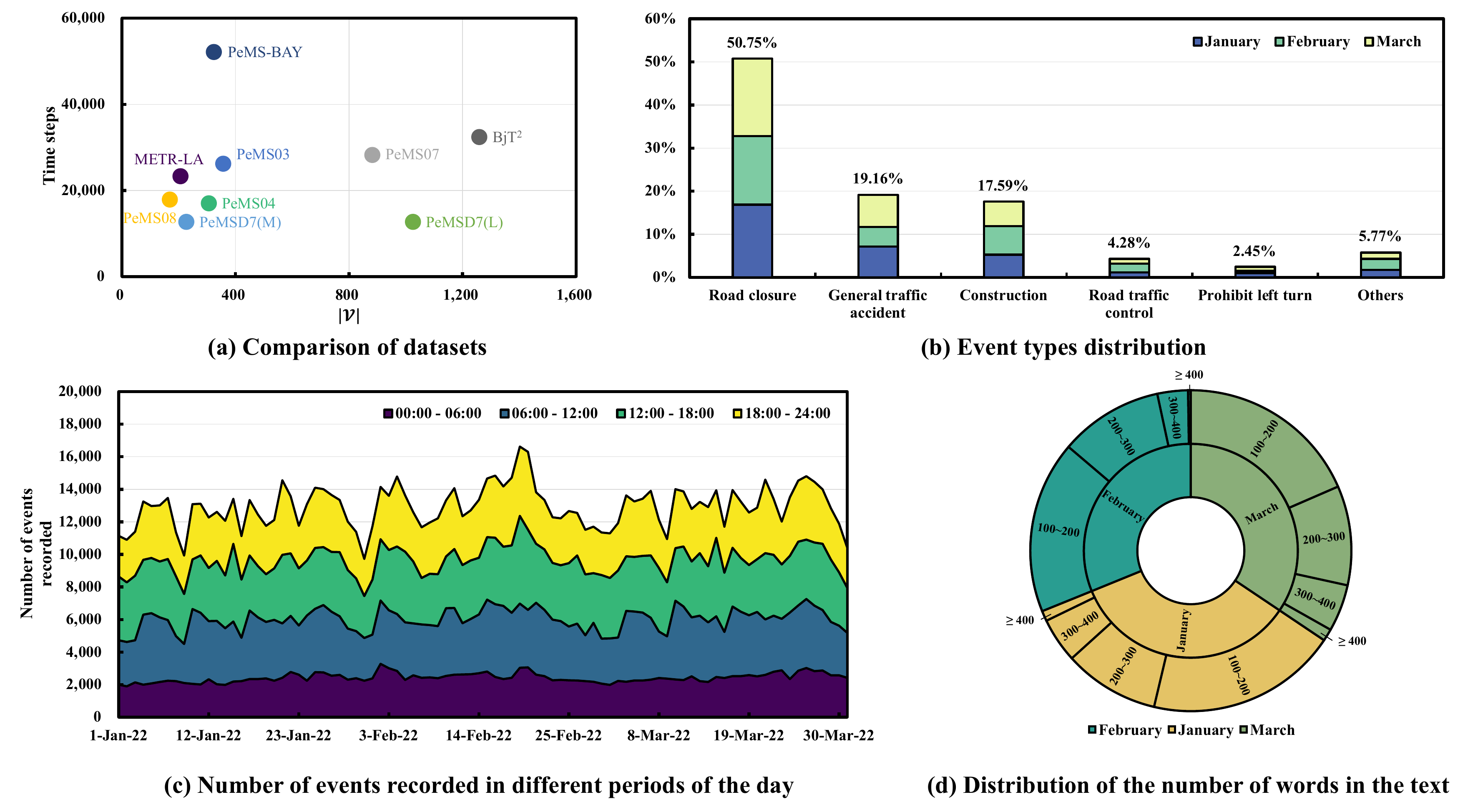}
   \caption{Dataset statistics of the BjTT dataset. (a) Comparison of datasets from the aspects of time steps and number of vertices, (b) proportion of top-frequent events out of all events, (c) number of events recorded in different periods of the day, (d) distribution of the number of words in the text.}
   \label{fig5}
\end{figure*}

\subsection{Data Statistics}
In this subsection, we analyze the properties of the BjTT dataset. The primary distinction between the BjTT dataset and other traffic prediction datasets lies in its inclusion of textual descriptions of events. Therefore, we predominantly concentrate on the analysis of the textual data.

\subsubsection{Data volume}
The BjTT dataset contains 1,260 road vertices as well as 32,400 time steps, spanning three months of traffic and event data. The comparison of the BjTT dataset with existing mainstream traffic prediction datasets from the aspects of both road vertices and time steps is illustrated in Figure \ref{fig5} (a). It can be observed that the PeMS-BAY dataset records the highest number of time steps at 52,116, while it ranks only fifth in terms of vertices. Contrarily, our BjTT dataset has the largest vertices number at 1,260 and secures the position of having the second-largest count of time steps at 32,400. Therefore, the BjTT dataset is highly competitive in data volume. In addition, as indicated in Table \ref{table1}, the BjTT dataset uniquely includes diverse types of traffic data, setting it apart from other datasets.

\subsubsection{Textual event data}
The BjTT dataset is notably characterized by its inclusion of textual data on various events, as depicted in Figure \ref{fig4}. This dataset features a wide range of event types, and Figure \ref{fig5} (b) demonstrates the distribution of these event types by percentage across the entire dataset. The five most common events identified are road closure, general traffic accident, construction, road traffic control, and prohibit left turn. It is worth noting that road closure accounts for over 50\% of all events, making it the most frequent type of event. The dataset does not delve into the specific causes behind each road closure, choosing instead to categorize them broadly of the road closure. This categorization likely contributes to the high frequency of road closure observed.

In addition, we conduct an analysis of event occurrences recorded during different periods of the day, as shown in Figure \ref{fig5} (c). We divide the whole day into four periods, namely 00:00 to 06:00, 06:00 to 12:00, 12:00 to 18:00, and 18:00 to 24:00. It can be seen that the period from 00:00 to 6:00 exhibits a comparatively lower number of recorded events when compared with other three periods. This disparity can primarily be attributed to the decreased frequency of emergencies such as traffic accidents during these early hours, though events like road construction may still exist. On a monthly scale, the trend in the number of events remained relatively consistent in January and March. However, the trend in February displays a notable distinction, particularly during the initial days of the month when the number of events drops off a cliff. The primary factor contributing to this phenomenon is the Chinese Spring Festival, which takes place from February 1st to February 6th, 2022.

Finally, we analyze the text lengths within the dataset, which comprises 32,400 texts. Text length plays a crucial role in the text encoding for various multimodal and language models \cite{28,29,30,31}. The distribution of text length in the BjTT dataset is illustrated in Figure \ref{fig5} (d). It can be seen that the majority of text data from January to March falls within the range of 100 to 200 words. There are also some texts with lengths ranging from 200 to 300 words, and a smaller portion exceeding 300 words. Such text lengths remain comfortably within the input length limit of most text encoders, making it possible to utilize the event text data provided by BjTT in specific tasks, thereby broadening the application of the data.

\section{Experiments and Analysis}
\label{section4}
In this section, we train six mainstream open-sourced traffic prediction methods on the proposed BjTT dataset, including STGCN\cite{3}, GWN\cite{5}, ASTGCN\cite{32}, STSGCN\cite{26}, MTGNN\cite{6}, STG-NCDE\cite{27}. Furthermore, to explore how textual data can assist the traffic prediction task, we involve a text-guided generative model in the comparison, which is named LDM\cite{33}. Finally, the experimental analysis and visualization results are discussed. It is worth noting that all experiments are conducted on the data type of velocity.

\subsection{Mainstream Methods Involved in Evaluation}
\textbf{STGCN \cite{3}:} Spatio-Temporal Graph Convolutional Networks. It is a classical and lightweight traffic prediction model, proposed by Yu \textit{et al.} in 2017. The authors formulate the traffic prediction problem on graphs instead of regular convolutional and recurrent units. This approach facilitates much faster training speed with fewer parameters.

\textbf{GWN \cite{5}:} Graph WaveNet. GWN effectively captures hidden spatial dependencies in the data by employing a novel adaptive dependency matrix, learned through node embedding. Besides, its stacked dilated 1D convolution component accommodates very long sequences, with the receptive field expanding exponentially with each additional layer.

\textbf{ASTGCN \cite{32}:} Attention based Spatial-Temporal Graph Convolutional Networks. ASTGCN mainly consists of three independent modules, each aimed at capturing a specific property of traffic flow: recent trends, daily periodic patterns, and weekly periodic dependencies. Every module contains two major parts: 1) a spatial-temporal attention mechanism; 2) a spatial-temporal convolution process. The outputs from these three modules are then combined through a weighted fusion process to produce the final prediction results.

\textbf{STSGCN \cite{26}:} Spatial-Temporal Synchronous Graph Convolutional Networks. This model effectively captures intricate localized spatial-temporal correlations by employing an elaborately designed spatial-temporal synchronous modeling mechanism. Additionally, it incorporates multiple modules tailored to different periods, enhancing its ability to capture the heterogeneities within localized spatial-temporal graphs.

\textbf{MTGNN \cite{6}:} Multivariate Time Graph Neural Networks. MTGNN is a general graph neural network tailored for multivariate time series data, proposed by Wu \textit{et al.} in 2020. This model automatically extracts the uni-directed relations among variables using a graph learning module, which seamlessly incorporates external knowledge such as variable attributes. Additionally, the authors introduce a novel mix-hop propagation layer and a dilated inception layer to effectively capture both spatial and temporal dependencies within the time series.

\textbf{STG-NCDE \cite{27}:} Spatio-Temporal Graph Neural Controlled Differential Equation. Neural-controlled differential equations represent a groundbreaking concept in the realm of sequential data processing. The authors extend the concept by designing two NCDEs, one dedicated to temporal processing and the other to spatial processing, which are subsequently integrated into a unified framework.

\textbf{LDM \cite{33}:} Latent Diffusion Models. LDM is a diffusion model implemented in the latent space of powerful pre-trained autoencoders, facilitating diffusion model training on limited computational resources without compromising quality or flexibility. By introducing cross-attention layers into the model architecture, the authors transform diffusion models into powerful and flexible generators capable of accommodating general conditioning inputs, such as text or bounding boxes, thereby enabling high-resolution synthesis in a convolutional manner.

\subsection{Evaluation Metrics}
Following some previous works, we adopt two metrics to evaluate the prediction performance, which are Mean Absolute Error (MAE) and Root Mean Squared Error (RMSE). These metrics are defined as follows:
\begin{equation}
    M A E=\frac{1}{N} \sum_{i=1}^{N}\left|x_{i}-\hat{x}_{i}\right|,
\end{equation}
\begin{equation}
    R M S E=\sqrt{\frac{1}{N} \sum_{i=1}^{N}\left|x_{i}-\hat{x}_{i}\right|^{2}},
\end{equation}
where MAE measures the average distance between the predicted value $\hat{x}_{i}$  and the actual value $x_{i}$, offering an overview of the global accuracy of the predicted traffic conditions. On the other hand, RMSE gives insight into the local quality of the generated traffic conditions, making it particularly responsive to outliers and noise. Here, $N$  denotes the number of predicted moments.

\begin{table*}[t]
\renewcommand{\arraystretch}{1.5}
  \centering
\caption{The performance comparison of several traffic prediction methods and generative models. These methods are evaluated on the test set of the three subdatasets of each month and the complete dataset. The best performance is highlighted in bold, while the second-best performance is underlined.}
  \resizebox{\textwidth}{!}
{
\begin{tabular}{c|c|c|cccccc|c}
\midrule[1pt]
                                      & $T$                      & Metric & STGCN & GWN  & ASTGCN & STSGCN & MTGNN & STG-NCDE & LDM \\ \midrule[1pt]
\multirow{6}{*}{BjTT (January)}       & \multirow{2}{*}{20 min} & MAE    & 3.80  & \underline{3.60} & 3.83   & 3.85   & \underline{3.60}  & 3.71     & \textbf{3.36}  \\
                                      &                        & RMSE   & 5.73  & \underline{5.51} & 5.82   & 5.89   & 5.53  & 5.75     & \textbf{5.18}   \\ \cline{2-10} 
                                      & \multirow{2}{*}{40 min} & MAE    & 4.24  & \underline{3.85} & 4.10   & 4.11   & 3.91  & 4.00     & \textbf{3.29}   \\
                                      &                        & RMSE   & 6.29  & \underline{5.91} & 6.39   & 6.40   & 5.92  & 6.26     & \textbf{5.05}   \\ \cline{2-10} 
                                      & \multirow{2}{*}{60 min} & MAE    & 4.55  & \underline{4.07} & 4.33   & 4.33   & 4.30  & 4.17     & \textbf{3.40}   \\
                                      &                        & RMSE   & 6.78  & \underline{6.10} & 6.74   & 6.76   & 6.26  & 6.59     & \textbf{5.22}   \\ \midrule[1pt]
\multirow{6}{*}{BjTT (February)}      & \multirow{2}{*}{20 min} & MAE    & 3.71  & 3.61 & 3.95   & 3.94      & 3.78  & \textbf{3.21}     & \underline{3.27}   \\
                                      &                        & RMSE   & 5.71  & 5.70 & 6.23   & 6.23      & 6.08  & \underline{5.51}     & \textbf{5.04}   \\ \cline{2-10} 
                                      & \multirow{2}{*}{40 min} & MAE    & 4.01  & 3.84 & 4.38   & 4.14      & 3.88  & \underline{3.51}     & \textbf{3.33}   \\
                                      &                        & RMSE   & 6.24  & 6.25 & 7.00   & 6.62      & 6.35  & \underline{6.20}     & \textbf{5.15}   \\ \cline{2-10} 
                                      & \multirow{2}{*}{60 min} & MAE    & 4.18  & 3.95 & 4.70   & 4.18      & 3.95  & \underline{3.72}     & \textbf{3.30}   \\
                                      &                        & RMSE   & 6.59  & \underline{6.43} & 7.55   & 6.66      & 6.47  & 6.67     & \textbf{5.11}   \\ \midrule[1pt]
\multirow{6}{*}{BjTT (March)}         & \multirow{2}{*}{20 min} & MAE    & 3.92  & \underline{3.51} & 3.77   & 3.60   & 3.58  & 3.56     & \textbf{3.23}   \\
                                      &                        & RMSE   & 5.97  & 5.52 & 5.85   & \underline{5.51}   & 5.55  & 5.65     & \textbf{4.98}   \\ \cline{2-10} 
                                      & \multirow{2}{*}{40 min} & MAE    & 4.55  & \underline{3.71} & 4.09   & 4.03   & 3.78  & 3.76     & \textbf{3.32}   \\
                                      &                        & RMSE   & 6.80  & 6.04 & 6.50   & 6.03   & \underline{5.96}  & 6.19     & \textbf{5.13}   \\ \cline{2-10} 
                                      & \multirow{2}{*}{60 min} & MAE    & 4.34  & \underline{3.83} & 4.28   & 4.17   & 4.25  & 3.89     & \textbf{3.25}   \\
                                      &                        & RMSE   & 6.87  & 6.26 & 6.86   & 6.53   & \underline{6.02}  & 6.42     & \textbf{5.10}   \\ \midrule[1pt]
\multirow{6}{*}{BjTT (Complete dataset)} & \multirow{2}{*}{20 min} & MAE    & \underline{3.56}  & 3.61 & 3.79   & 3.66      & 3.60  & \underline{3.56}     & \textbf{3.30}   \\
                                      &                        & RMSE   & \underline{5.45}  & 5.60 & 5.87   & 5.74      & 5.60  & 5.53     & \textbf{5.08}   \\ \cline{2-10} 
                                      & \multirow{2}{*}{40 min} & MAE    & 3.85  & 3.92 & 4.15   & 4.01      & \underline{3.74}  & 3.83     & \textbf{3.26}   \\
                                      &                        & RMSE   & 5.93  & 6.19 & 6.54   & 6.22      & \underline{5.91}  & 6.09     & \textbf{5.07}   \\ \cline{2-10} 
                                      & \multirow{2}{*}{60 min} & MAE    & 4.00  & 4.16 & 4.44   & 4.15      & \underline{3.89}  & 4.05     & \textbf{3.24}   \\
                                      &                        & RMSE   & \underline{6.19}  & 6.61 & 7.02   & 6.63      & \textbf{6.19}  & 6.52     & \textbf{5.00}   \\ \midrule[1pt]
\end{tabular}
}
  \label{table2}
\end{table*}

\begin{figure*}[h!]
  \centering
   \includegraphics[width=\textwidth]{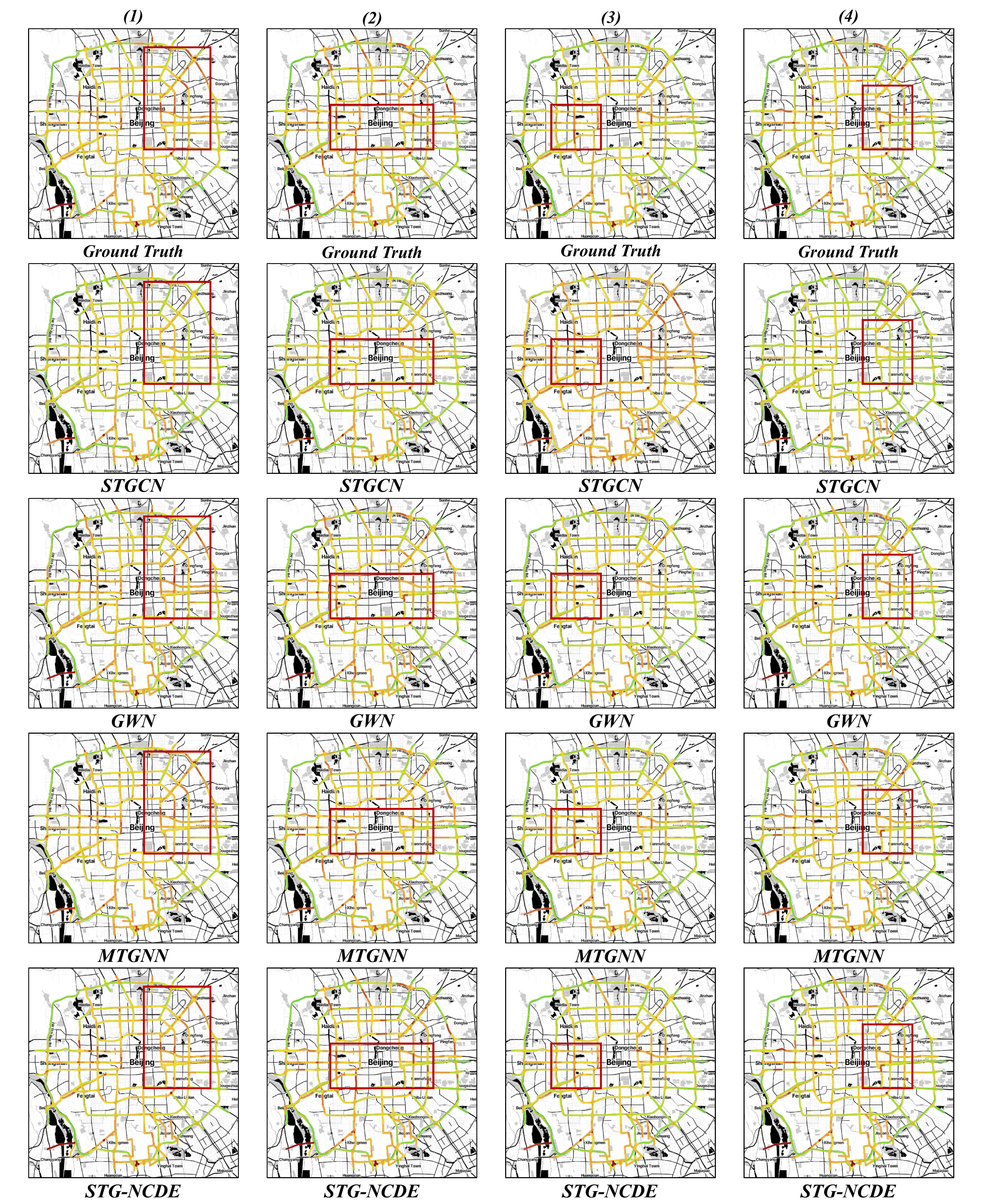}
   \caption{The four groups of qualitative comparison of some selected methods (STGCN, GWN, MTGNN, and STG-NCDE). The first to fourth columns represent four specific junctures. Best viewed in red boxes.}
   \label{fig6}
\end{figure*}

\begin{figure*}[h!]
  \centering
   \includegraphics[width=\textwidth]{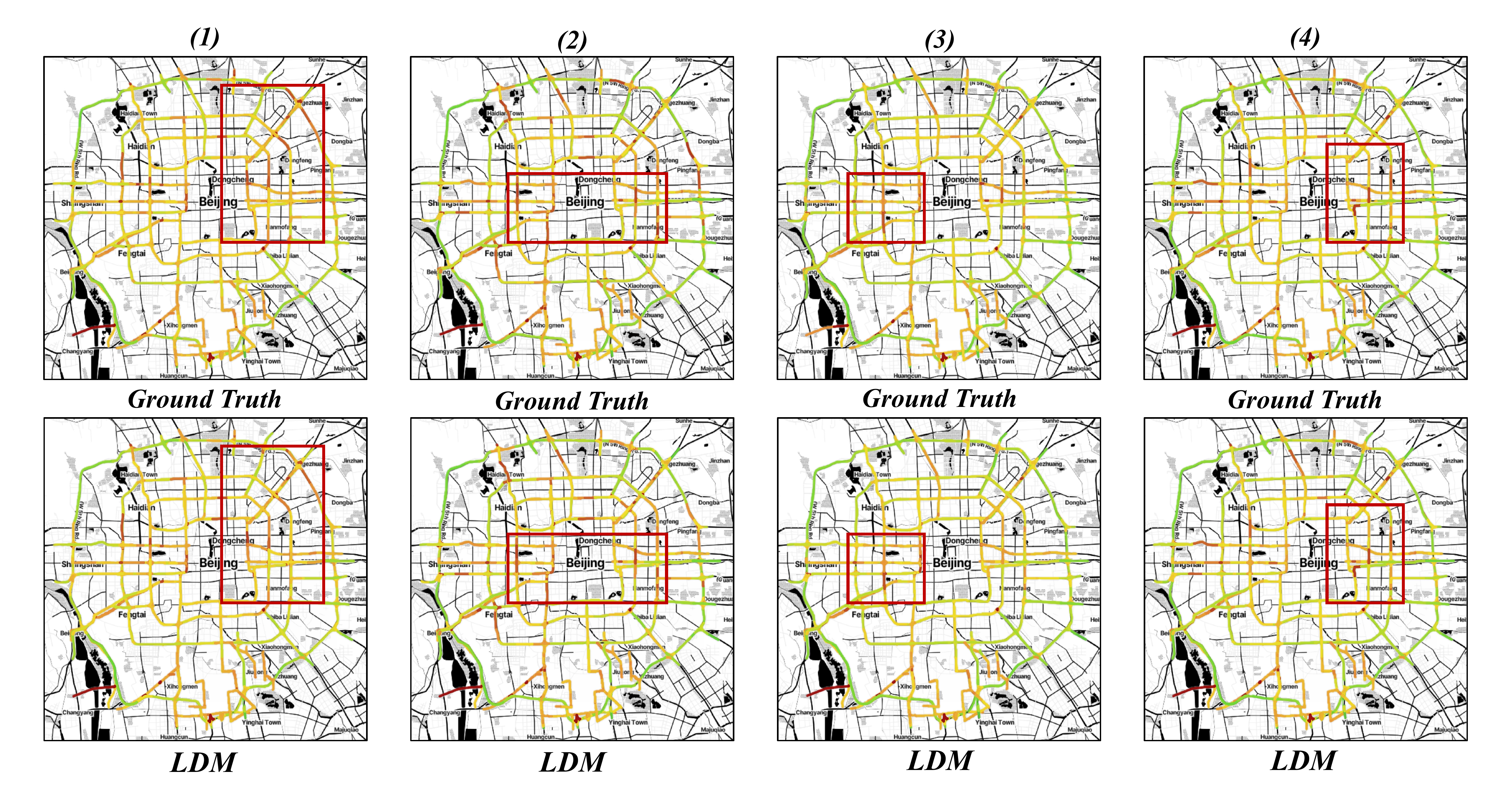}
   \caption{The four groups of qualitative comparison of LDM and ground truth. The first to fourth columns represent four specific junctures. Best viewed in red boxes.}
   \label{fig7}
\end{figure*}

\begin{figure}[h!]
  \centering
   \includegraphics[width=1.\columnwidth]{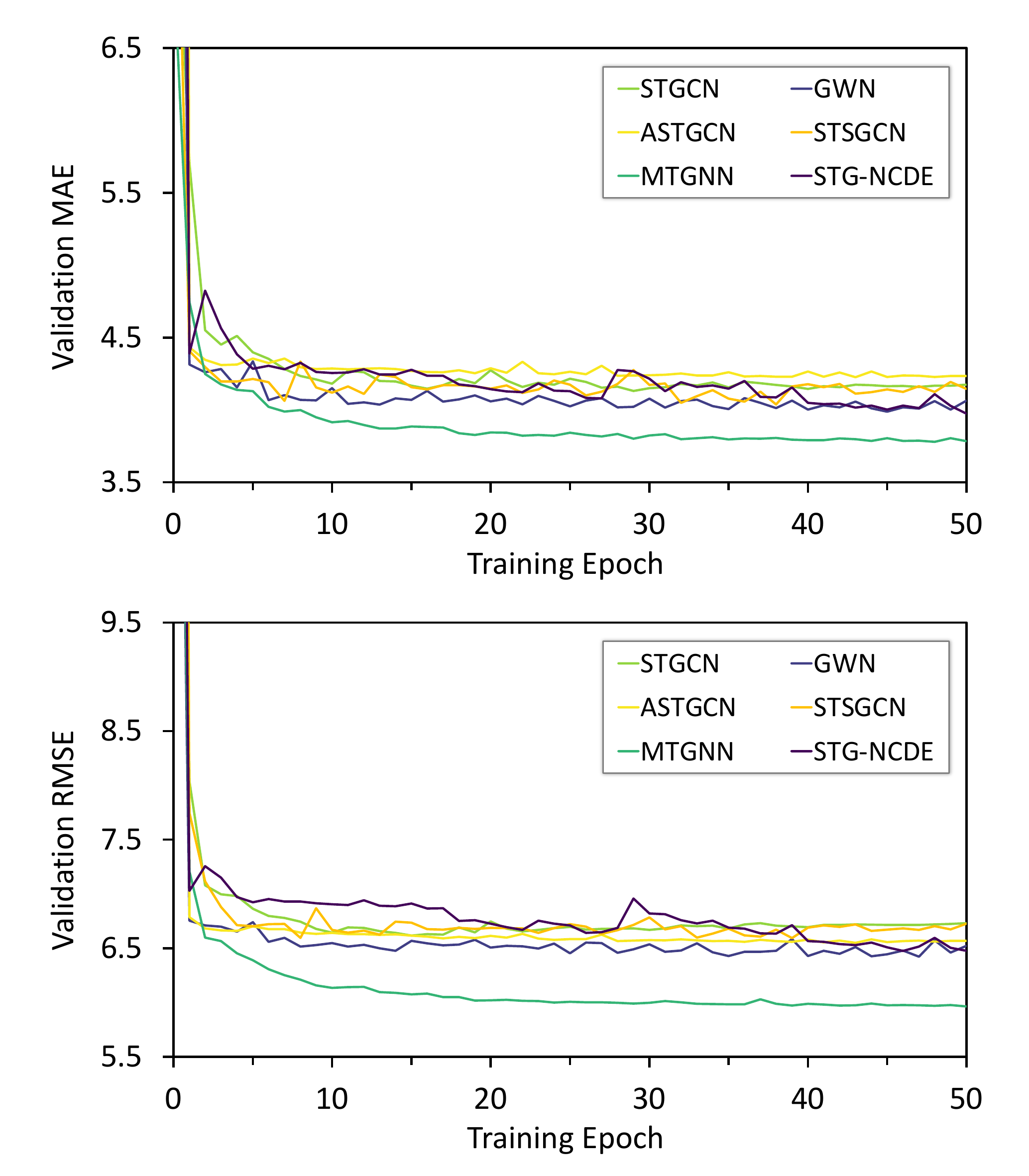}
   \caption{Validation MAE versus the training epoch (up). Validation RMSE versus training epochs (down). The training is conducted on the complete BjTT dataset for the prediction of subsequent 15 data points.}
   \label{fig8}
\end{figure}

\subsection{Implementation Details}
The experiment mainly comprises two primary parts: 1) training on data from three distinct months within the BjTT dataset, and 2) training on the complete BjTT dataset. For all the traffic prediction models in the experiment, the models are trained based on the official codes and the default parameters. We customize the input configurations of these models to enable the simultaneous prediction of time-series data across 1,260 vertices. The dataset is partitioned into a training set, validation set, and test set in proportions of 70\%, 10\%, and 20\%, respectively. For the text-guided generative model, we reshape the input traffic data $x_{i}$ to $x_{i} \in \mathbb{R}^{36 \times 36 \times 2}$ and keep the rest of the parameters unchanged. All the experiments are conducted on a single Nvidia GeForce 3090Ti ($\sim$ 24GB).

\subsection{Results Analysis}
\subsubsection{Quantitative results}
Table \ref{table2} presents the performance comparison of several methods evaluated on the BjTT dataset. Each method is trained and tested using data from independent months in the dataset as well as the complete dataset. For traffic prediction methods, GWN and STG-NCDE consistently emerge as top contenders, showing superior performance on monthly data. On the other hand, STGCN and MTGNN demonstrate better performance over the complete dataset, indicating their adaptability to diverse data patterns and large data volumes. A consistent observation across all traffic prediction methods is the inverse relationship between prediction time range and accuracy, highlighting a common challenge in time-series methods where extended prediction intervals tend to degrade performance.

The BjTT dataset is notably characterized by its inclusion of event data. Therefore, we simultaneously evaluate a text-guided generation model named the latent diffusion model on the BjTT dataset for directly generating traffic situations corresponding to events. Diverging from time-series prediction methods, the LDM uniquely operates on textual input during the inference stage. To present this feature, the test set for the LDM is further divided into three subsets in chronological order. The performance metrics for each subset are calculated independently and are listed in the 20, 40, and 60-minute prediction rows of the time-series prediction methods. Contrary to the time-dependent degradation observed in traditional traffic prediction methods, the performance of LDM remains robust across varying prediction intervals. In this way, the application potential of such text-guided generative models is substantial in real-world contexts. For instance, to predict the traffic impact of large social events like concerts or sports matches, these models enable the preemptive generation of traffic situations, thereby facilitating the formulation of contingency strategies to alleviate congestion.

\subsubsection{Qualitative results}
Figure \ref{fig6} illustrates some visualization results of predicted traffic situations by four traffic prediction methods. The top row displays the ground truth at four specific junctures, while the subsequent rows, from the second to the fifth, depict the predictions outputted by different methods. The results of all traffic prediction methods are based on the former 15 time steps (1 hour) of the target junctures to ensure the highest prediction accuracy. It can be seen that all four junctures have different levels of congestion in different areas, as shown by the red boxes in the figure. Both MTGNN and STG-NCDE  successfully capture most of the congested zones, whereas STGCN shows a less accurate correspondence with the ground truth. This demonstrates that MTGNN and STG-NCDE have a better capability in short-term traffic prediction.

Figure \ref{fig7} presents the traffic situations generated by the LDM with the input of textual descriptions of events. The generative results from LDM closely align with the actual congested zones. Compared with traditional traffic prediction methods, the generative model bypasses the necessity of historical data during inference, enabling it to generate traffic situations for any future junctures unrestrained by temporal limitations.

\subsubsection{Training efficiency}
To further explore the performance of the traffic prediction methods, we plot the MAE and RMSE across the validation set as they are being trained on the BjTT dataset, as shown in Figure \ref{fig8}. Observations on the speed of convergence indicate that the methods are largely on par with one another. However, MTGNN distinguishes itself by achieving the lowest MAE and RMSE on the validation set, followed closely by GWN and STG-NCDE. This finding aligns with the test set results listed in Table \ref{table2}. Broadly speaking, the performance of these traffic prediction methods on the BjTT dataset is basically similar to their performance on traditional datasets such as the PeMS series, affirming the BjTT dataset's applicability and relevance for traffic prediction tasks.

\section{Conclusion and Outlooks}
\label{section5}
In this paper, the first large-scale public multimodal dataset BjTT for traffic prediction is constructed, which has the characteristics of large data volume, diverse data type, and new data modality. It contains 1,260 road vertices, 32,400 time steps, and 3 different traffic data types. Based on the proposed dataset, we evaluate six state-of-the-art traffic prediction methods and an outstanding text-guided generative model, and give a quantitative and qualitative analysis of the results. 

The analysis highlights two principal challenges in the domain of traffic prediction that the proposed dataset helps to mitigate. 1) Insensitive to abnormal events. Urban transport networks often encounter abnormal events, such as vehicular accidents, roadworks, or severe weather conditions, which can significantly impact traffic situations. Methods that rely solely on historical data tend to falter in accurately predicting traffic situations under these abnormal circumstances. For example, as indicated in Table \ref{table2}, STGCN fails to anticipate congestion effectively. 2) Limited performance in long-term prediction. Long-term prediction plays a significant role in traffic management. As evidenced in Table \ref{table1}, there exists substantial scope for improvement in long-term traffic prediction accuracy. 

Based on the experimental results, the generative model with text data provided by  BjTT as input significantly alleviates the aforementioned two challenges. In the future, we will continue to focus on exploring the potential of generative models for improving the performance of traffic prediction.

{
\bibliographystyle{IEEEtran}
\bibliography{main}
}

\begin{IEEEbiography}[{\includegraphics[width=1in,height=1.25in,clip,keepaspectratio]{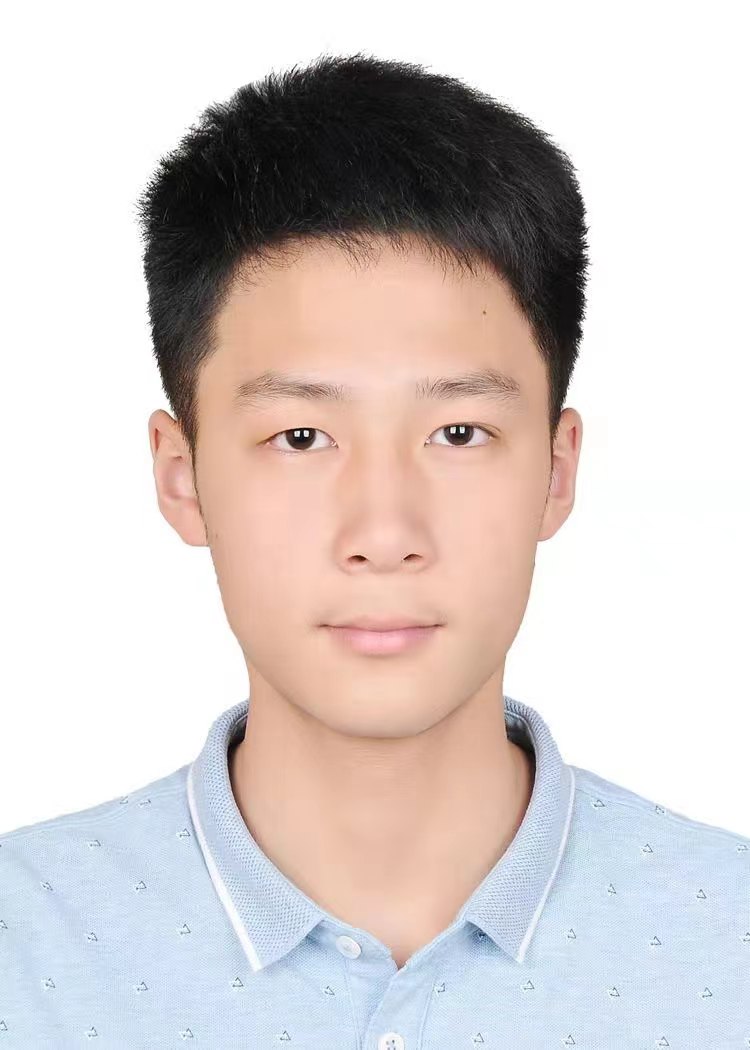}}]{Chengyang Zhang} received his bachelor degrees from Beijing Information Science and Technology University. He is currently a graduate student at the Department of Informatics, Beijing University of Technology. His research interests include computer vision, traffic prediction and biomedical image analysis.
\end{IEEEbiography}

\begin{IEEEbiography}[{\includegraphics[width=1in,height=1.25in,clip,keepaspectratio]{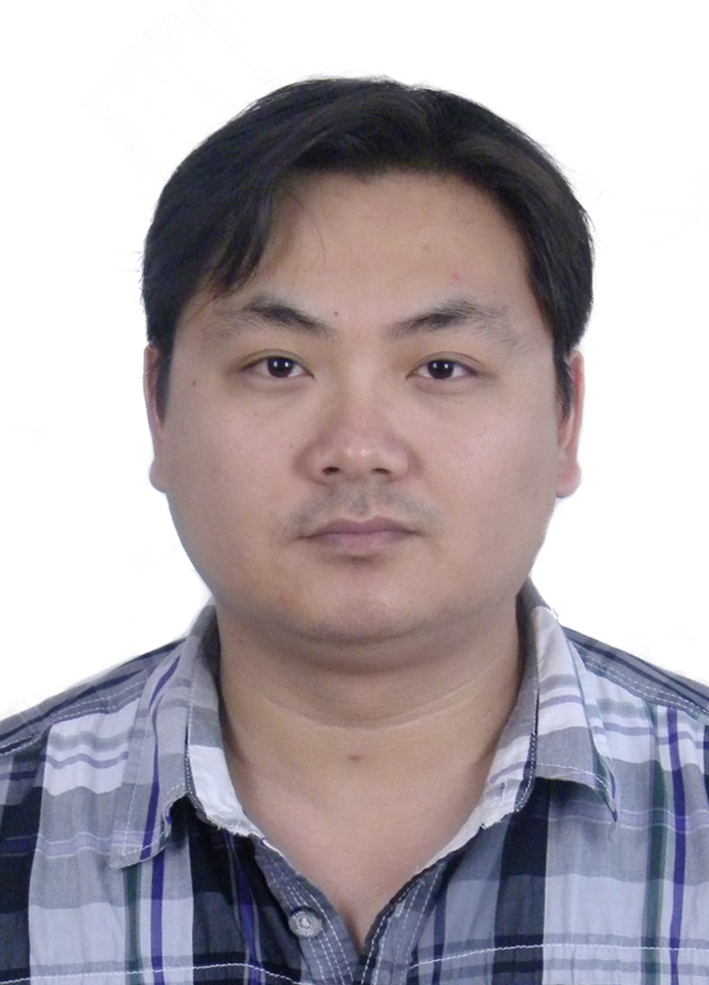}}]{Yong Zhang} (Member, IEEE) received the Ph.D.degree in computer science from the Beijing University of Technology in 2010. He is currently an Associate Professor in computer science with the Beijing University of Technology. His research interests include intelligent transportation systems, big data analysis, visualization, and computer graphics.
\end{IEEEbiography}

\begin{IEEEbiography}[{\includegraphics[width=1in,height=1.25in,clip,keepaspectratio]{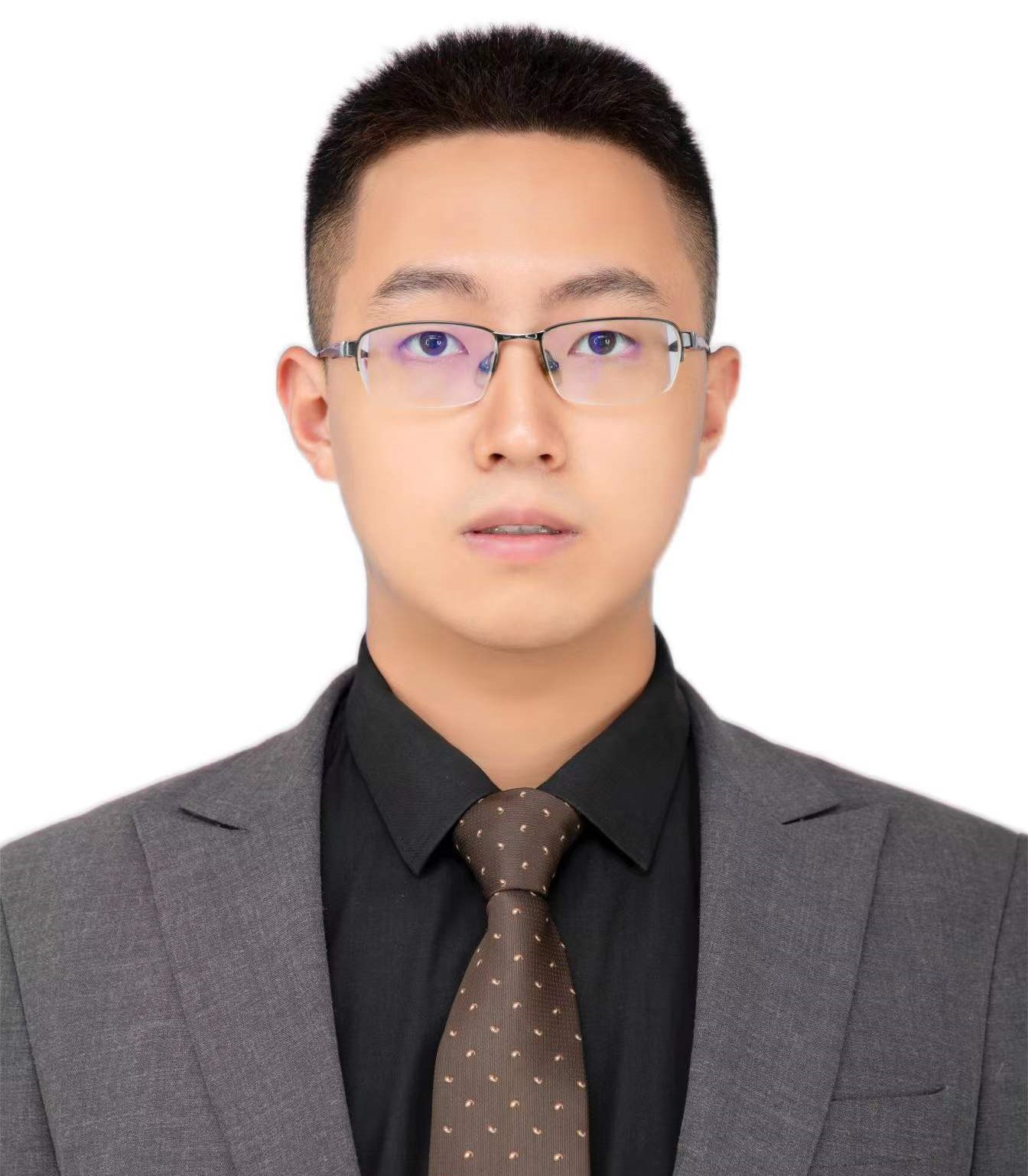}}]{Qitan Shao} received the bachelor’s degree in computer science and technology from the Beijing University of Technology. He is currently a graduate student at the Department of Informatics, Beijing University of Technology. His research interest is intelligent transportation systems.
\end{IEEEbiography}

\begin{IEEEbiography}[{\includegraphics[width=1in,height=1.25in,clip,keepaspectratio]{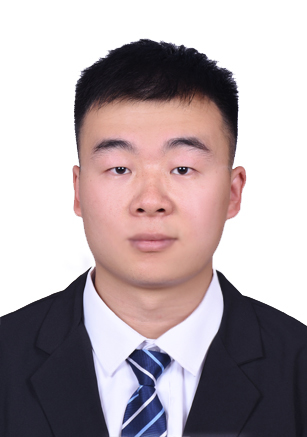}}]{Jiangtao Feng} is currently pursuing a B.S. degree in Electronic Information from Beijing University of Technology, China. His research includes traffic anomaly detection, prediction and spatio-temporal data mining.
\end{IEEEbiography}

\begin{IEEEbiography}[{\includegraphics[width=1in,height=1.25in,clip,keepaspectratio]{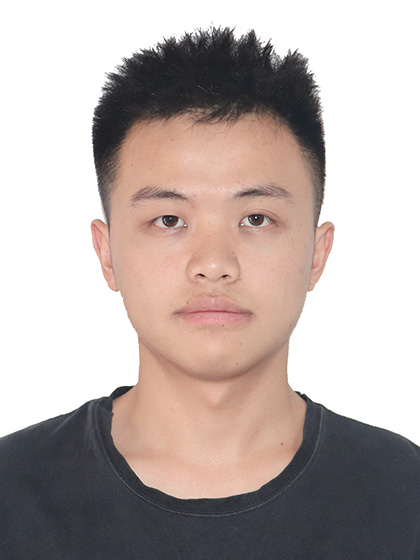}}]{Bo Li} received his bachelor degrees from Beijing Information Science and Technology University. He is currently a graduate student at the Department of Informatics, Beijing University of Technology. His research interests include computer vision and biomedical image analysis.
\end{IEEEbiography}

\begin{IEEEbiography}[{\includegraphics[width=1in,height=1.25in,clip,keepaspectratio]{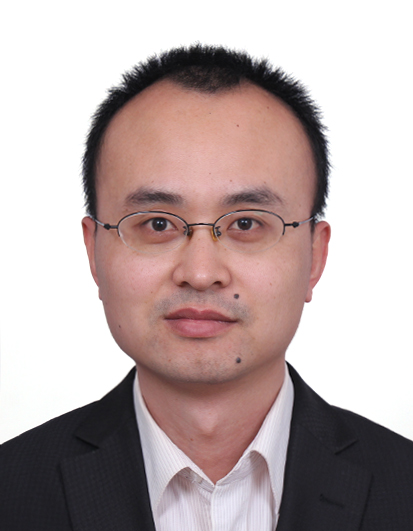}}]{Yisheng Lv} received the B.E. and M.E. degrees in transportation engineering from Harbin Institute of Technology, Harbin, China, in 2005 and 2007, respectively, and the Ph.D. degree in control theory and control engineering from Chinese Academy of Sciences, Beijing, China, in 2010. He is an Assistant Professor with State Key Laboratory of Management and Control for Complex Systems, Institute of Automation, Chinese Academy of Sciences. His research interests include traffic data analysis, dynamic traffic modeling, and parallel traffic management and control systems.
\end{IEEEbiography}

\begin{IEEEbiography}[{\includegraphics[width=1in,height=1.25in,clip,keepaspectratio]{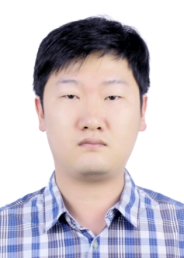}}]{Xinglin Piao} received the Ph.D. degree from the Beijing University of Technology, Beijing, China, in 2017. He is currently a lecturer in the Faculty of Information Technology at Beijing University of Technology. His research interests include intelligent traffic, pattern recognition, and multimedia technology.
\end{IEEEbiography}


\begin{IEEEbiography}[{\includegraphics[width=1in,height=1.25in,clip,keepaspectratio]{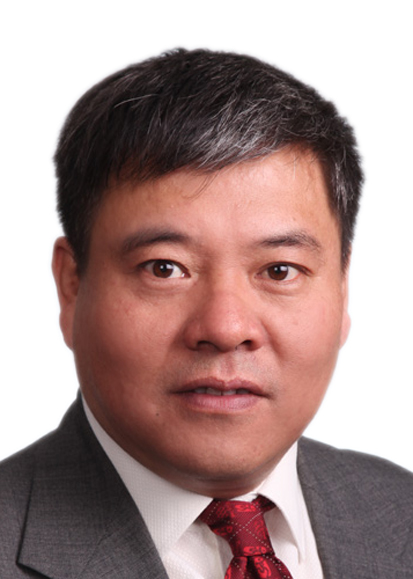}}]{Baocai Yin} (Member, IEEE) received the B.S., M.S., and Ph.D. degrees in computational mathematics from the Dalian University of Technology, Dalian, China, in 1985, 1988, and 1993, respectively. He is currently a Professor with the Beijing Key
Laboratory of Multimedia and Intelligent Software Technology, Faculty of Information Technology, Beijing University of Technology. His research interests include multimedia, image processing, computer vision, and pattern recognition.
\end{IEEEbiography}

\end{document}